%% file: paper.tex
\documentclass[sigconf]{acmart}

\usepackage{balance}

\usepackage{fp}
\usepackage{hyperref}
\usepackage{amsmath}
\usepackage{makecell}
\usepackage{multirow}
\usepackage{multicol}
\usepackage{comment}
\usepackage{xspace}
\usepackage{tikz}
\usetikzlibrary{calc}
\usepackage{pgf}
\usepackage{xcolor}
\usepackage{colortbl}
\usepackage{pifont}
\usepackage[most]{tcolorbox}
\usepackage{graphicx}
\usepackage{tabularx}
\usepackage{subcaption}
\usepackage{ifthen}
\usepackage{csvsimple}
\usepackage{siunitx}
\usepackage{etoolbox}
\usepackage{pgffor}
\usepackage{forloop}
\usepackage{arydshln}
\usepackage{makecell}
\usepackage{url}
\usepackage{hhline}
\usepackage{booktabs}
\usepackage{amssymb}
\usepackage{amsthm}

\theoremstyle{plain}
\newtheorem{thm}{\protect\theoremname}
\theoremstyle{plain}
\newtheorem{prop}[thm]{\protect\propositionname}
\theoremstyle{remark}

\theoremstyle{definition}
\newtheorem{defn}[thm]{\protect\definitionname}

\providecommand{\definitionname}{Definition}
\providecommand{\propositionname}{Proposition}
\providecommand{\remarkname}{Remark}
\providecommand{\theoremname}{Theorem}

\usetikzlibrary{arrows,shadows,shapes,positioning,calc}

\newcommand{\mypara}[1]{\vspace*{0.06in}\noindent\textbf{#1}\xspace}

\newcommand*{\eg}{\textit{e.g.}\@\xspace}

\newcommand*{\etal}{\textit{et al.}\@\xspace}

\newcommand{\RS}[1]{RS-#1\xspace}

\newcommand*{\dropout}{Dropout\xspace}
\newcommand*{\randomspiking}{Random Spiking\xspace}

\newcommand*{\WHTBOX}{\text{white-box}\@\xspace}

\newcommand*{\GRYBOX}{\text{translucent-box}\@\xspace}
\newcommand*{\capitalGRYBOX}{\text{Translucent-box}\@\xspace}
\newcommand*{\BLKBOX}{\text{black-box}\@\xspace}

\newcommand*{\CW}{C\&W\@\xspace}

\newcommand*{\MNIST}{\text{MNIST}\@\xspace}
\newcommand*{\FASHION}{\text{Fashion-MNIST}\@\xspace}
\newcommand*{\CIFAR}{\text{CIFAR-10}\@\xspace}

\makeatletter
\newcommand*{\etc}{%
    \@ifnextchar{.}%
    {\textit{etc}}%
    {\textit{etc}.\@\xspace}%
}

\newcommand\Tstrut{\rule{0pt}{2.6ex}}         
\newcommand\Bstrut{\rule[-1.0ex]{0pt}{0pt}}   

\def\BibTeX{{\rm B\kern-.05em{\sc i\kern-.025em b}\kern-.08emT\kern-.1667em\lower.7ex\hbox{E}\kern-.125emX}}
    
\copyrightyear{2020}
\acmYear{2020}
\setcopyright{acmcopyright}
\acmConference[CODASPY '20]{Proceedings of the Tenth ACM Conference on Data and Application Security and Privacy}{March 16--18, 2020}{New Orleans, LA, USA}
\acmBooktitle{Proceedings of the Tenth ACM Conference on Data and Application Security and Privacy (CODASPY '20), March 16--18, 2020, New Orleans, LA, USA}
\acmPrice{15.00}
\acmDOI{10.1145/3374664.3375736}
\acmISBN{978-1-4503-7107-0/20/03}

\begin{document}

\fancyhead{}

\title{Random Spiking and Systematic Evaluation of Defenses Against Adversarial Examples}

\author[H Ge]{Huangyi Ge}
\email{geh@purdue.edu}
\affiliation{%
    \institution{Purdue University}
}
\author[SY Chau]{Sze Yiu Chau}
\email{schau@purdue.edu}
\affiliation{%
    \institution{Purdue University}
}
\author[B. Ribeiro]{Bruno Ribeiro}
\email{ribeiro@cs.purdue.edu}
\affiliation{%
    \institution{Purdue University}
}
\author[N. Li]{Ninghui Li}
\email{ninghui@cs.purdue.edu}
\affiliation{%
    \institution{Purdue University}
}

%

%
\begin{abstract}
    Image classifiers often suffer from adversarial examples, which are generated by strategically adding a small amount of noise to 
    input images to trick classifiers into misclassification. Over the years, many defense mechanisms have been proposed,
    and different researchers have made seemingly contradictory claims on their effectiveness. 
    We present an analysis of possible adversarial models, and propose an evaluation framework for comparing different defense mechanisms. 
    As part of the framework, we introduce a more powerful and realistic adversary strategy.
    Furthermore, we propose a new defense mechanism called Random Spiking (RS), which generalizes dropout and 
    introduces random noises in the training process in a controlled manner. 
    Evaluations under our proposed framework suggest RS delivers better protection against adversarial examples than many existing schemes.
\end{abstract}

%
%


%

\sloppypar
%
\maketitle

\section{Introduction}
\input{src/1_intro}

\section{Background}
\input{src/2_background}

\section{Evaluation Methodology}
\input{src/3_threat}

\section{Proposed Defense}
\input{src/4_implementation}

\section{Experimental Evaluation}
\input{src/5_rspike_training}

\section{Related Work}
\input{src/6_related}

\section{Conclusion}
\input{src/7_conclusion}

%
\bibliographystyle{ACM-Reference-Format}
\bibliography{dl}

%
\appendix
\input{src/appendices}

\end{document}

%% file: src/1_intro.tex
\label{sec:intro}

Modern society increasingly relies on software systems trained by machine 
learning (ML) techniques.  Many such techniques, however, were designed under the 
implicit assumption that both the training and test data follow the same static 
(although possibly unknown) distribution.  In the presence of intelligent and 
resourceful adversaries, this assumption no longer holds.  
Such an adversary can deliberately manipulate a test instance, and cause the trained models to behave unexpectedly.
For example, it is found that existing image classifiers based on Deep Neural
Networks are highly vulnerable to adversarial examples~\cite{szegedy_ZSBEGF_13,
goodfellow_fgd_15}.  Often times, by modifying an image in a way that is barely
noticeable by humans, the classifier will confidently classify it as something else.
This phenomenon also exists for classifiers that do not use neural networks, and has
been called ``optical illusions for machines''.
Understanding why adversarial examples work and how to defend against them is
becoming increasingly important, as machine learning techniques are used
ubiquitously, for example, in transformative technologies such as autonomous cars, 
unmanned aerial vehicles, and so on.

Many approaches have been proposed to help defend against adversarial
examples.  Goodfellow et al.~\cite{szegedy_ZSBEGF_13} proposed
adversarial training, in which one trains a neural network using both the original
training dataset and the newly generated adversarial examples.  In region-based
classification~\cite{cao_G_2017}, one aggregates predictions on multiple perturbed 
versions of an input instance to make the final prediction. 
Some approaches attempt to train additional neural network models to identify and 
reject adversarial examples~\cite{mengccs17, xu_EQ_2018}.

We point out that since the adversary can choose instances and shift the test distribution 
\emph{after} a model is trained, adversary examples exist so long as ML models differ from 
human perception on some instances.  (These instances can be used as adversarial examples.)  
Thus adversarial examples are unlikely to be completely eliminated.  What we can do is to reduce the number of such instances by training ML models that better match human perceptions, and by making it more difficult for the attacker to find adversarial examples.  

While the research community has seen a proliferation in proposals of defense mechanisms, conducting a thorough evaluation and a fair head-to-head comparison of different mechanisms remains challenging.  
In Section~\ref{sec:systematic-eval}, we analyze possible adversarial models, and propose to conduct evaluation in a variety of models, including both \WHTBOX and \GRYBOX attacks.  In \GRYBOX attacks, the adversary is assumed to know the defense mechanism, model architecture, and distribution of training data, but not the precise parameters of the target model.  With this knowledge, the adversary can train one or more \textbf{surrogate models}, and to generate adversarial examples leveraging such surrogate models.  

While other research efforts have attempted to generate adversarial examples based on surrogate models and then assess 
transferability, existing application of this method does not fully exploit the potential of 
surrogate models.  As a result, one can overestimate the effectiveness of defenses.  We propose two 
improvements.  First, one can train many surrogate models under the same configuration, and 
then generate adversarial  examples that can simultaneously fool multiple surrogate models at the same time.
Second, one can reserve some surrogate models as
``validation models''.  These validation models are not used when generating adversarial 
examples; however, generated adversarial examples are first run against them, and only those 
examples that are able to fool a certain percentage of the validation models are used
in evaluation against the target model.  This models a more determined and resourceful attacker
who is willing to spend more resources to find more effective adversarial examples to deploy, 
a scenario that is certainly realistic.  Our experimental results demonstrate that these more
sophisticated adversary strategies lead to significantly higher transferability rates. 

Furthermore, in Section~\ref{sec:implementation}, we propose a new defense 
mechanism called Random Spiking, where random noises are added to one or more hidden 
layers during training.  Random Spiking generalizes the idea of dropout~\cite{srivastava_HKSS_2014}, 
where hidden units are randomly dropped during training.  In Random Spiking, the 
outputs of some randomly chosen units are replaced by a random noise during training.  

In Section~\ref{sect:eval}, we present extensive evaluations of several existing defense
mechanisms and Random Spiking (RS) under both \WHTBOX and \GRYBOX attacks, and
empirically show that RS, especially when combined with adversarial training, improve the
resiliency against adversarial examples.

In summary, we make three contributions.  (1) The proposed evaluation methodology, 
especially the more powerful and realistic adversary strategy of attacking multiple surrogates 
in parallel and using validation models to filter.  (2) The idea of Random Spiking, which 
is demonstrated to offer additional resistance to adversarial examples.  
(3) We provide a thorough evaluation of several defense mechanisms against 
adversarial examples, improving our understanding of them.

%% file: src/2_background.tex
\label{sec:background}

We consider neural networks that are used as $m$-class classifiers, where the output
of a network is computed using the softmax function.
Given such a neural network used for classification, let $\mathbf{z}(x)$ denote the vector output of the final layer before the softmax activation,
and $C(x)$ denote the classifier defined by the neural network.
Then,
\\[-2pt]
\centerline{$C(x) = \underset{i}{\arg \max} \Big(
\exp(\mathbf{z}(x)_i)\Big/\big(\sum_{j=1}^{n}\exp(\mathbf{z}(x)_j)\big)\Big).$}
\\[0pt]
Oftentimes, $\exp(\mathbf{z}(x)_i)\big/\big(\sum_{j=1}^{n}\exp(\mathbf{z}(x)_j\big)$
is interpreted as the probability that the input $x$ belongs to the $i$-th category, and
the classifier chooses the class with the highest probability. Under this interpretation,
the output $\mathbf{z}(x)$ is related to log of odds ratios, and is thus called the \textbf{logits output}.

\subsection{Adversarial Examples}
\label{sec:adv_expl}

Given a dataset $D$ of instances, each of the form $(x,y)$, where $x$ gives the features of the instance, and $y$ 
the label, and a classifier $C(\cdot)$ trained using a subset of $D$, we say that an instance $x'$ 
is an \emph{adversarial example} if and only if \textbf{there exists an instance $(x,y) \in D$ such that $x'$ is 
    close to $x$, $C(x)=y$, and $C(x') \ne y$}.

Note that in the above we did not define what ``$x'$ is close to $x$'' means.  Intuitively, when $x$ represents an image, by closeness we mean human perceptual similarity.  However, we are unaware of any mathematical distance metric that accurately measures human perceptual similarity.  In the literature $L_p$ norms are often used as the distance metric for closeness.  $L_p$ is defined as\\[0pt]
\centerline{$L_p(x, x') = \|x-x'\|_p = \Big(\sum_{i=1}^n|{x}_i-{x'}_i|^p\Big)^{1/p}.$} \\[0pt]
The commonly used $L_p$ metrics include: $L_0$, the number of changed pixels \cite{papernot_MJFCS_15}; $L_1$, the 
sum of absolute values of the changes in all pixels \cite{chen_SZYH_2017};
$L_2$, the Euclidean norm \cite{carlini_oakland_17, moosavi_Dezfooli_15, szegedy_ZSBEGF_13,
carlini_magnet_attack17}; and $L_{\infty}$, the maximum absolute change \cite{goodfellow_fgd_15}.
In this paper, we use $L_2$, which reflects both the number of changed pixels and the magnitude of their change.  We 
call $L_2(x,x')$ the \textbf{distortion} of the adversarial example $x'$.

When generating an adversarial example against a classifier $C(\cdot)$, one typically starts from an existing instance 
$(x,y)$ and generates $x'$.  In an \textbf{untargeted attack}, one generates $x'$ such that $C(x') \ne y$.  In a 
\textbf{targeted attack}, one has a desired target label $t\ne y$ and generates $x'$ such that $C(x')=t$.

Goodfellow \etal \cite{goodfellow_fgd_15} proposed the fast gradient sign (\textbf{FGS}) attack, which generates adversarial examples based on the gradient sign of the loss value according  to the input image.  A more effective attack is proposed by Carlini and 
Wagner~\cite{carlini_oakland_17}, which we call the \textbf{\CW} attack. 
Given a neural network with logits output $\mathbf{z}$, an input $x$, and a target class label 
$t$, the \CW attack tries to solve the following optimization problem:
\vspace{-0.3em}
\begin{equation}\label{eqn:attack_obj}
 \underset{x'}{\arg \min} \left(\|x-x'\|_p + c \cdot l(x') \right)
 \vspace{-0.3em}
\end{equation}
where the loss function $l$ is defined as
$$ l(x') = \max\left(\max\left\{\mathbf{z}(x')_i: i\neq t\right\} - \mathbf{z}(x')_t,
-K\right).
$$
Here, $K$ is called the \textbf{confidence value}, and is a positive number that one can choose.
Intuitively, we desire
$\mathbf{z}(x')_t$ to be higher than any $\mathbf{z}(x')_i$ where $i\neq t$ so that
the neural network predicts label $t$ on input $x'$.  Furthermore, we prefer the gap in
the logit of the class $t$ and the highest of any class other than $t$ to be as large as
possible (until the gap is $K$, at which point we consider the gap to be sufficiently
large).  
In general, choosing a large value $K$ would result in adversarial examples that have a higher distortion, but will be classified to
the desired label with higher confidence.
The parameter $c>0$ in Eq.~(\ref{eqn:attack_obj}) is a regularization constant to adjust the relative importance of 
minimizing the distortion versus minimizing the loss function $l$.
In the attack, $c$ is initially set to a small initial value, and then dynamically adjusted based on the progress made by the iterative optimization process. 

The \CW attack uses the Adam algorithm~\cite{Kingma_B_2014} to solve the optimization problem in Eq.~(\ref{eqn:attack_obj}).  Adam performs iterative gradient-based optimization, based on adaptive estimates of 
lower-order moments.
Compared to other attacks, such as FGS, the \CW attack is more time-consuming.  However, it is able to  find more effective adversarial examples.

\subsection{Existing Defenses}
\label{sec:defenses}

Many approaches have been proposed to help defend against adversarial examples.
Here we give an overview of some of them.

\mypara{Adversarial Training.}
Goodfellow \etal~\cite{goodfellow_fgd_15} proposed to train a neural network using both the training dataset and newly generated adversarial examples.
In \cite{goodfellow_fgd_15}, it is shown that models that have gone through adversarial training provide some resistance against adversarial examples generated by the FGS method.

\mypara{Defensive Distillation.}
Distillation training was originally proposed by Hinton  
\etal~\cite{hinton_vd_distill_2015} for the purpose of distilling knowledge out of a 
large model (one with many parameters) to train a more compact model (one with 
fewer parameters).  Given a model whose knowledge one wants to distill, one 
applies the model to each instance in the training dataset and generates a probability 
vector, which is used as the new label for the instance.  This is called the soft label, 
because, instead of a single class, the label includes probabilities for different 
classes.  A new model is trained using instances with soft labels.  The intuition is that 
the probabilities, even those that are not the largest in a vector, encode valuable 
knowledge.  To make this knowledge more pronounced, the probability vector is 
generated after dividing the logits output with a temperature constant $T>1$.  This 
has the effect of making the smaller probabilities larger and more pronounced.  The 
new model is trained with the same temperature.  However, when deploying the 
model for prediction, temperature is set to $1$.

\emph{Defensive Distillation} \cite{papernot_MWJS_15} is motivated by the original
distillation training proposed by Hinton \etal \cite{hinton_vd_distill_2015}.  The main
difference between the two training methods is that defensive distillation uses the
same network architecture for both initial network and distilled network.  This is because
the goal of using Distillation here is not to train a model that has a smaller size, but to
train a more robust model.

\mypara{\dropout.}
\dropout \cite{srivastava_HKSS_2014} was introduced to improve generalization accuracy through the
introduction of randomness in training. 
The term ``dropout'' refers to dropping out units, i.e., temporarily removing the units along with all its incoming and outgoing connections. 
In the simplest case, during each training epoch, each unit is retained with a fixed probability $p$ independent of other units, where $p$ can be chosen using a validation
set or can simply be set to 0.5, which was suggested by the authors 
of~\cite{srivastava_HKSS_2014}.

There are several intuitions why \dropout is effective in reducing generalization errors.  One is that
after applying \dropout, the model is always trained with a subset of the
units in the neural network.  This prevents units from co-adapting too much. That is,
a unit cannot depend on the existence of another unit, and needs to learn
to do something useful on its own.  Another intuition is that training
with \dropout approximates simultaneous training of an exponential number
of ``thinned'' networks.  In the original proposal, dropout is applied in training, but not in testing.  
During testing, without applying \dropout, the prediction approximates an averaging output of all these thinned networks.  In Monte Carlo dropout~\cite{gal_g_2016}, dropout is also applied in testing. The NN is run multiple times, and the resulting prediction probabilities are averaged for making prediction.  This more directly approximates the behavior of using the NN as an ensemble of models. 

Since \dropout introduces randomness in the training process, two models that are trained with \dropout are likely to be less similar than
two models that are trained without using \dropout.  Defensive 
Dropout~\cite{wang_wzwkcl_2018} explicitly uses dropout for defense against 
adversarial examples.  It applies dropout in testing, but runs the network just once.  In 
addition, it tunes the dropout rate used in testing by iteratively generating 
adversarial examples and choosing a drop rate to both maximize the testing 
accuracy and minimize the attack success rate.

\mypara{Region-based Classification.}
Cao and Gong \cite{cao_G_2017} proposed region-based classification to defend against adversarial examples. Given an 
input, region-based classification first generates $m$ perturbed inputs by adding bounded random noises to the 
original input, then computes a prediction for each perturbed input, and finally use voting to make the final 
prediction.  This method slows down prediction by a factor of $m$. In~\cite{cao_G_2017}, $m=10,000$ was used for MNIST and $m=1,000$ was used for CIFAR.  Evaluation in \cite{cao_G_2017} shows that this can withstand adversarial examples generated by the \CW attack under low confidence value $K$.  However, if one slightly increases the confidence value $K$ when generating the adversarial examples, this defense is no longer effective.
\vspace{0em}

\mypara{MagNet.}
Meng and Chen~\cite{mengccs17} proposed an approach that is called MagNet.  MagNet combines two ideas to defend
against adversarial examples.  First, one trains detectors that attempt to detect and reject adversarial
examples.  Each detector uses an autoencoder, which is trained to minimize the $L_2$ distance between input image and
output.  A threshold is then selected using validation dataset.  The detector rejects any image such that the $L_2$
distance between it and the encoded image is above the threshold.  Multiple detectors can be used.
Second, for each image that passes the detectors, a reformer (another autoencoder) is applied to the image, and
the output (reformed image) is sent to the classifier for classification.

The evaluation of MagNet in~\cite{mengccs17} considers only adversarial examples generated without knowledge of the MagNet defence.   
Since one can 
combine all involved neural networks into a single one, one can still apply the \CW attack on the composite network.  In~\cite{carlini_magnet_attack17}, an effective attack is carried 
out against MagNet by adding to the optimization objective a term describing the goal of evading the detectors.

\vspace{-0.25em}

%% file: src/3_threat.tex
\label{sec:systematic-eval}

We discuss several important factors for evaluation, and introduce the \GRYBOX model to supplement \WHTBOX evaluation.
\vspace{-0.75em}

\subsection{Adversary Knowledge}

Adversary model plays an important role in any security evaluations.  
One important part of the adversary model is the assumption on 
adversary's knowledge. 

\mypara{Knowledge of Model (\textbf{\WHTBOX}).}
    The adversary has full knowledge of the target model to be attacked, 
    including the model architecture, defense mechanism, and all the 
    parameters, including those used in the defense mechanisms. We call 
    such an attack a \WHTBOX attack.

\mypara{Complete Knowledge of Process (\GRYBOX).}
    The adversary does not know the exact parameters of the target model, but knows the training process, including model architecture, defense mechanism, training algorithm, and distribution of the training dataset. 
    With this knowledge, the adversary can use the same training process that is used to generate the target model 
    to train one or more \textbf{surrogate models}.
    Depending on the degree of randomness involved in the training process, the surrogate models may be similar
    or quite different from the target model, and adversarial examples generated by attacking
    the surrogate model(s) may or may not work very well. The property of whether an
    adversarial example generated by attacking one or more surrogate models can also work against
    another target model is known as \textbf{transferability}. We call such an attack a  \GRYBOX attack.

    Technically, it is possible for a \WHTBOX adversary to know less than a \GRYBOX adversary \emph{in some aspects}.  For example, a \WHTBOX adversary may not know the distribution of the training data.  However, for the purpose of generating adversarial examples, knowing all details of the target model (\WHTBOX) is strictly more powerful than knowing the training process (\GRYBOX).

\mypara{Oracle access only (\BLKBOX).}
Some researchers have considered adversary models where an adversary uses only oracle accesses to the target model.  That is, the adversary may be able to query the target model with instances and receive the output.  This is also called ``decision-based adversarial attack''~\cite{brendel_rb_2018, chen_j_2019}.  We call such an attack a \BLKBOX attack.

Some researchers also use \BLKBOX attack to refer to what we call \GRYBOX attacks. 
We choose to distinguish \GRYBOX attacks from \BLKBOX attacks for two reasons.  
First, an adversary will have some knowledge about the target model under attack, e.g., the neural network architecture and 
the training algorithm.  Thus the box is not really ``black''.  
Second, the two kinds of attacks are very different.  One relies on training surrogate models, and the other relies on issuing a large number of oracle queries. 

Other researchers use ``\BLKBOX attack'' to refer to the situation that the adversary carries out the attack without specifically targeting the defense mechanism. 
We argue that such an evaluation has limited values in understanding the security benefits of a defense 
mechanism, as it is a clear deviation from the Kerckhoffs's principle.

\mypara{Our Choice of Adversary Model.}
We argue that defense mechanisms should be evaluated under both \WHTBOX and \GRYBOX attacks.  While developing attacks that can generate adversarial examples using only oracle access is interesting, for a defense mechanism to be effective, one must assume that the adversary cannot break it even if it has the knowledge of the defense mechanism.   

Evaluation under \WHTBOX attack can be carried out by measuring the level of distortion needed to attack a model.  
Effective defense against \WHTBOX attacks is the ultimate objective.  
Until defense in the \WHTBOX model is achieved, effective defense against \GRYBOX attacks is valuable and help the research community make progress.
\capitalGRYBOX is a realistic assumption especially in an academic setting, as published papers generally include
descriptions of the architecture, training process, defense mechanisms and the exact
dataset used in their experiments. Robustness and security evaluations under this
assumption is also consistent with the Kerckhoffs's
principle. 

We also note that there are two possible flavors of attacks. Focusing on image
classifiers, the goal of an \emph{untargeted attack} is to generate adversarial
examples such that the classifier would give any output labels different from what
human perception would classify. A \emph{targeted attack} would additionally require the
working adversarial examples to induce the classifier into giving specific output
labels of the attacker's choosing.  In this paper, we consider only targeted
attacks when evaluating defense mechanisms, as it models an adversary with a 
more specific objective.

\subsection{Adversary Strategy}
\label{sect:attacker-strategy}

Even after the assumption about the adversary's knowledge is made, 
there are still possibilities regarding what strategy the adversary takes.
For example, when evaluating a defense mechanism under the \GRYBOX 
assumption, a standard method is to train $m$ models, and, for each model, 
generate $n$ adversarial examples.  Then for each of the $m$ model, 
treat it as the target model, and feed the $(m-1)n$ adversarial examples generated
on other models to it, and report the percentage of success among the $m(m-1)n$ trials.

Such an evaluation method is assessing the success probability of the 
following naive adversary strategy: The adversary trains one surrogate model, 
generates an adversarial example that works against the surrogate model, and 
then deploy that adversarial example.  We call this a \textbf{one-surrogate attack}. 
A real adversary, however, can use 
a more effective strategy.  It can try to generate adversarial examples 
that can fool multiple surrogate models at the same time.  After generating 
them, it can first test whether the adversarial examples can fool surrogate 
models that are not used in the generation.  We call this a \textbf{multi-surrogate attack}.

For any defense mechanism that is more effective against adversarial examples 
under the \GRYBOX attack than under the \WHTBOX model, the additional 
effectiveness must be due to the randomness in the training process. When 
that is the case, the above adversary strategy would have much higher success 
rate than the naive adversary strategy.  Evaluation should be done against 
this adversary strategy.

We thus propose the following procedure for evaluating a defense mechanism 
in the \GRYBOX setting.  One first trains $t+v$ surrogate models.  Then a set of 
$t$ models are randomly selected, and adversarial examples are generated 
that can \emph{simultaneously} attack all $t$ of them; that is, the optimization 
objective of the attack includes all $t$ models.  For the remaining $v$
models, we use leave-one-out validation.  That is, for each model, we 
use $v-1$ model as validation models, and select only adversarial examples 
that can fool a certain fraction of the validation models.  Only for the examples 
that pass this validation stage, do we record whether it successfully transfer to the target 
model or not.  We call such an attack a \textbf{multi-surrogate with validation attack}. 
The percentage of the successful transfer is used for evaluation. 
In our experiments, we use $t=v=8$, and an example is selected when it can successfully attack at least 
5 out of 7 validation models. 

\subsection{Parameters and Data Interpretation}

Training a defense mechanism often requires multiple parameters as inputs.  For
example, a defense mechanism may be tuned to be more vigilant against
adversarial examples, at the cost of reduced classification accuracy.  When
comparing defense mechanisms, one should choose parameters 
in a way that the classification accuracy on test dataset is similar.

At the same time, when using the \CW attack to generate adversarial examples, an important parameter is the confidence value $K$.  A defense mechanism may be able to resist adversarial examples generated under a low $K$ value, but may prove much less effective against those generated under a higher value $K$ (see, e.g.,~\cite{cao_G_2017}).  
Using the same $K$ value for different defenses, however, may not be sufficient for providing a level playing field for comparison.  The $K$ value represents an input to the algorithm, and what really matters is the quality of the adversarial examples.  We propose to run the \CW attack against a defense mechanism under multiple $K$ values, and group the resulting adversarial examples based on their distortion.  We can then compare how well a defense mechanism performs against adversarial examples with similar amount of distortion.  That is, we group adversarial examples based on the $L_2$ distance and compute 
the average transferability for each group.

%% file: src/4_implementation.tex
\label{sec:implementation}

From a statistical point of view, the problem with adversarial examples is that of 
classification under covariate shifts~\cite{shimodaira2000improving}.  A covariate 
shift happens when the training and test observations follow different distributions.  
In the case of adversarial examples, this is clearly the case, as new adversarial 
examples are generated and added to the test distribution.  
If the test distribution with adversarial examples can be known, a simple and optimal way for dealing with covariate shifts is training the model with samples from the test distribution, rather than using the original training data~\cite{shimodaira2000improving,sugiyama2017dataset,sugiyama2008direct,heckman1977sample}, assuming that we have access to enough such examples.  Training with adversarial examples can be viewed as a robust optimization procedure~\cite{madry2017towards} approximating this approach. 
 
Unfortunately, training with adversarial examples  does not fully solve the defense problem. 
Adversaries can adapt the test distribution (a new covariate shift) to make the new classifier perform poorly again on test data.  That is, given a model trained with adversarial examples, the adversary can find additional adversarial examples and use them. 
In this minimax game, where the adversary is looking for a covariate shift and the defender is training with the latest covariate shift, the odds are stacked against the defender, who is always one step behind the attacker~\cite{tsipras2018robustness,zhang2019theoretically}.  

Fundamentally, to win this game, the defender needs to mimic human perception. That is, as long as there are instances (real or fabricated) where humans and ML models classify differently, these can be over represented in the test data by the adversary's covariate shift. 
Models that either underfit or overfit both make mistakes by definition, and these 
mistakes can be used in the adversary's covariate shift.  Only a model with no 
training or generalization errors under all covariate shifts is not vulnerable to 
attacks.

\subsection{Motivation of Our Approach}

While it is impossible to completely eliminate classification errors, several things can be done to help defend against adversarial examples by making them harder to find.  

One approach is to reduce the number of instances that the ML models disagree with human perception.  Training with adversarial examples help in this regard.  Using more robust model architecture and training procedure can also help.  
When giving an image to train the model, intuitively we want to say that ``all instances that look similar to this instance from a human's perspective should also have the same label''.  Unfortunately, finding which images humans will consider to be ``similar to this instace'' and thus should be of the same class is not a well-defined procedure.  
Today, the best we can hope for is that for some mathematical distance measure (such as $L_2$ distance) and with a smaller enough threshold, humans will consider the images to be similar.  If we substitute ``look similar to ... from a human's perspective'' with ``within a certain $L_2$ distance'', this is a precise statement.  This suggests that one training instance should be interpreted as a set of instances (e.g., those within a certain $L_2$ distance of the given one) all have the same given label.  Our proposed defense is to some extent motivated by this intuition. 

Another way is to make it more difficult for adversaries to discover adversarial 
examples, even if they exist.  One approach is to use an ensemble model, wherein 
multiple models are trained and applied to an instance and the results are 
aggregated in some fashion.  For an adversarial example to work, it must be able to 
fool a majority of the models in the ensemble.  

If we consider defense in the translucent box adversary model, another approach is to increase the degree of randomness in the training process, so that adversarial examples generated on the surrogate models do not transfer well. 

Our proposed new defense against adversarial examples are motivated by these 
ideas, which are recapped below.  First, each training instance should be viewed as 
representatives of instances within a certain $L_2$ distance.  Second, we want to 
increase the degree of randomness in the training process.  Third, we want to 
approximate the usage of an ensemble of models for decision.  

\subsection{\randomspiking}

As discussed in Section~\ref{sec:defenses}, dropout has been proposed as a way to defend against adversarial examples.  
Dropout can be interpreted as a way of regularizing a neural network by adding noise to its hidden units. 
The idea of adding noise to the states of units has also been used in
the context of Denoising Autoencoders (DAEs) by Vincent et al.~\cite{Vincent_lb_2008, Vincent_kkibm_2010}, where noise is added to the input units of an autoencoder and the network is trained to reconstruct the noise-free input.  \dropout changes the behavior of the hidden units. 
Furthermore, instead of adding random noises, in \dropout, values are set to zero. 

Our proposed approach generalizes both \dropout and Denoising Autoencoders.
Instead of training with removed units or injecting random noises into
the input units, we inject random activations into some hidden units near the 
input level.  We call this method \textbf{Random Spiking}.  Similar to dropout, there are two approaches 
at inference time.  The first is to use random spiking only in training, and does not use 
it at inference time.  The second is to use a Monte Carlo decision procedure.  
That is, at decision time, one runs the NN multiple times with random spiking, and aggregate
the result into one decision. 

The motivations for random spiking are many-fold. 
First, we are simulating the interpretation that each training instance should 
be treated as a set of instances, each with some small changes.  Injecting 
random perturbations at a level near the input simulates the effect of 
training with a set of instances.  
Second, adversarial examples make only small perturbations on benign images that
do not significantly affect human perception. These perturbations inject
noises that will be amplified through multiple layers and change the
prediction of the networks.  Random Spiking trains the network to be
more robust to such noises.  Third, if one needs to increase the degree
of randomness in the training process beyond \dropout, using random
noises instead of setting activations to zero is a natural approach.  
Fourth, when we use the Monte Carlo decision procedure, we are approximating the behavior of 
a model ensemble.  

More specifically, random spiking adds a filtering layer in between two layers of nodes in a DNN.  
The effect of the filtering layer may change the output values of 
units in the earlier layer, affecting the values going into the later layer.  
With probability $p$, a unit's value is kept unchanged. With 
probability $1-p$, a unit's value is set to a randomly sampled noise.  
If a unit has its output value thus randomly perturbed, in
back-propagation we do not propagate backward through this unit, 
since any gradient computed is related to the random noise, and not the actual behavior of this unit. 
For layers after the \randomspiking filtering layer, back-propagation update would occur normally.

We use the \randomspiking filtering layer just once, after the first convolutional layer (and before
any max pooling layer if one is used).  This is justified by the design intuition.  We also 
experimented with adding the  \randomspiking filtering layer later in the NN, and test accuracy 
drops. 
There are two explanations for that.  First, since units chosen to have random noises stop 
back-propagation, having them later 
in the network has more impact on training.  Second, when random noises are injected 
early in the network, there are more layers after it, and there is sufficient capacity 
in the model to deal with such noises without too much accuracy cost. When random noises 
are injected late, fewer layers exist to deal with their effect, and the network lacks the capacity 
to do so.  

\mypara{Generating Random Noises.}
To implement \randomspiking, we have to decide how to sample
the noises that are to be used to replace the unit outputs.
Sampling from a distribution with a fixed range is problematic
because the impact of noise depends on the distribution of other values in the same layer.  
If a random perturbation is too small compared to other values in the same layer, 
then its randomization effect is too small. If, on the other hand, the magnitude of the noise is 
significantly larger than the other values, it overwhelms the network.  
In our approach, we compute the minimum and maximum value among all values in the layers 
to be filtered, and sample a value uniformly at random in that range.  
Since training NN is often done using mini-batches, the minimum and maximum values are computed
from the whole batch.

\mypara{Monte Carlo Random Spiking as a Model Ensemble.}
For testing, we can use the Monte Carlo decision procedure of running the network multiple times and use
the average.  This has attractive theoretical guarantees, at the cost of overhead for decision time, since
the NN needs to be computed multiple times for one instance.  
We now show that the Monte Carlo Random Spiking approximates a model ensemble.  
Let $(x,y)$ be a training example, where $x$ is an image and $y$
is the image's one-hot encoded label. Consider a RS neural network
with softmax output $\hat{y}(x,\boldsymbol{b},\boldsymbol{\epsilon},\boldsymbol{W})$,
neuron weights $\boldsymbol{W}$, and spike parameters $\boldsymbol{b}$
and $\boldsymbol{\epsilon}$, where bit vector $\boldsymbol{b}_{i}=1$
indicates that the $i$-th hidden neuron of the RS layer gives out
a noise output $\boldsymbol{\epsilon}_{i}\in\mathbb{R}$ sampled with
density $f(\epsilon)$, otherwise $\boldsymbol{b}_{i}=0$ and the
output of the RS layer is a copy of its $i$-th input from the previous layer (i.e., the original value of the neuron). 
By construction, $\boldsymbol{b}_{i}=1$ with
probability $1-p$ independent of other RS neurons. Let $L(y,\hat{y})$
be a convex loss function over $\hat{y}$, such as the cross-entropy
loss, the negative log-likelihood, or the square error loss. Then,
the following proposition holds:

\begin{prop}
Consider the ensemble RS model \label{def:RSprop}
\vspace{-0.3em}
\begin{equation}
\overline{\hat{y}}(x,\boldsymbol{W})\equiv\sum_{\forall\boldsymbol{b}}\int_{\boldsymbol{\epsilon}}\hat{y}(x,\boldsymbol{b},\boldsymbol{\epsilon},\boldsymbol{W})p(\boldsymbol{b})f(\boldsymbol{\epsilon})d\boldsymbol{\epsilon},\label{eq:yprime}
\end{equation}
where $f$ is a density function, $p(\boldsymbol{b})$ is the probability
that bit vector $\boldsymbol{b}$ is sampled, and $\hat{y}$ is a
RS neural network with one spike layer. Then, by stochastically optimizing
the original RS neural network $\hat{y}$ by sampling bit vectors
and noises, we are performing the minimization
\[
\boldsymbol{W}^{\star}=\arg\!\min_{\boldsymbol{W}}L(y,\overline{\hat{y}}(x,\boldsymbol{W}))
\]
through a variational approximation model using an upper bound of the loss $L(y,\overline{\hat{y}}(x,\boldsymbol{W}))$.
A proof of Prop.~\ref{def:RSprop} is presented in Appendix~\ref{app:proof}.
\end{prop}
\begin{defn}[MC Avg. Inference] \label{def:MCavg}
At inference time, we use Monte Carlo sampling to estimate the RS
ensemble
\vspace{-0.3em}
\[
\overline{\hat{y}}(x,\boldsymbol{W})=\sum_{\forall\boldsymbol{b}}\int_{\boldsymbol{\epsilon}}\hat{y}(x,\boldsymbol{b},\boldsymbol{\epsilon},\boldsymbol{W})p(\boldsymbol{b})f(\boldsymbol{\epsilon})d\boldsymbol{\epsilon,}
\vspace{-0.3em}
\]
where $f$ is a density function, $p(\boldsymbol{b})$ is the probability
that bit vector $\boldsymbol{b}$ is sampled.
\end{defn}
\vspace{-1.3em}

\mypara{Adaptive Attack against \randomspiking.}
Since \randomspiking introduces randomness during training, an adaptive attacker knowing that 
\randomspiking has been deployed but is unaware of the exact parameters 
of the target model can train multiple surrogate models, and try to generate adversarial examples that
can simultaneously cause all these models to misbehave.  
That is, the multi-surrogate with validation is a natural adaptive attack against \randomspiking, and 
any other defense mechanisms that rely on randomness during training.  
In this attack, one uses probabilities from all surrogate models to generate the adversarial example.  
This is similar to the Expectation over Transformation (EOT)~\cite{athalye_elik_2018} approach for
generating adversarial examples. 

%% file: src/5_rspike_training.tex
\label{sect:eval}

\input{src/table_dataset_info.tex}

We present experimental results comparing the various defense mechanisms using 
our proposed approach. 
\vspace{-0.8em}

\subsection{Dataset and Model Training}
\label{eval-subsect:baseline}

For our experiments, we use the following 3 datasets:
{\MNIST}~\cite{lecun_cc_1998},
{\FASHION}~\cite{xiao_rv_2017}, and
{\CIFAR}~\cite{krizhevsky_09}. Table~\ref{table:dataset-info} gives an overview of their characteristics.

We consider $9$ schemes equipped with different defense mechanisms, all of which share the 
the same network architectures and training parameters. 
For \MNIST, we follow the architecture given in the \CW paper~\cite{carlini_oakland_17}. 
\FASHION was not studied in the literature in an adversarial setting, and 
the model architectures used for \CIFAR in previous papers delivered a fairly low 
accuracy.  
Thus for \FASHION and \CIFAR, we use the state-of-the-art WRN-28-10 instantiation of the wide residual networks~\cite{zagoruyko_K_16}.
We are able to achieve state-of-the-art test accuracy using these architectures.  
Some of these mechanisms have adjustable parameters, and we choose values for these parameters so that the resulting models have a comparable level of accuracy on the testing data.  As the result, all $9$ schemes result in small accuracy drop.

Table~\ref{table:model-acc-full} gives the test errors, and Tables \ref{table:model-arch} and  \ref{table:model-para} in the Appendix give details of the model architecture, and training parameters. 
When a scheme uses either Dropout or Random Spiking, we consider 
3 possible decision procedures at test time.  By ``Single pred.'', we mean dropout and 
random spiking are not used at test time.  By ``Voting'', we mean running the 
network with Dropout and/or Random Spiking 10 times, and use  majority voting for 
decision (with ties decided in favor of the label with smaller index).  By ``MC Avg.'', 
we mean using Definition~\ref{def:MCavg} by running the network with Dropout 
and/or Random Spiking 10 times, and averaging the 10 probability vectors.  For each 
scheme, we train $16$ models (with different initial parameter values) on each 
dataset, and report the mean and standard deviation of their test accuracy. 
We observe that using Voting or MC Avg, one can typically achieve a slight reduction 
in test error. 

\subsubsection{Adversarial training}

\label{eval-subsect:adv-examples}

Two defense mechanisms require training with adversarial examples, which are generated by applying the \CW $L_2$ 
targeted attack on a target model, using randomly sampled training instances and target class labels. 

\subsubsection{Upper Bounds on Perturbation.}
For each dataset, we generated thousands of adversarial examples with varying confidence values for 
each training scheme, and have them sorted according 
to the added amount of perturbation, measured in $L_2$. 
We have observed that from instances that have high amount of perturbation one can visually observe 
the intention of adversarial example.  We thus chose a cut-off upper bound on $L_2$ distance.  
The chosen $L_2$ cut-off bounds are included in Table~\ref{table:para-adv}, and used as upper limits in many of our later experiments. 
With the bounds on $L_2$ fixed, we then empirically determine an upper bound for the confidence value to be used in 
the \CW-$L_2$ attacks for generating adversarial examples for training purposes. 
To diversify the set of generated adversarial examples, we sample several different confidence values within the bound, which are also reported 
in Table~\ref{table:para-adv}. 
\input{src/test-error-table-full-data}
\begin{table}
    \centering
    \caption{Parameters used for generating adversarial examples. 
    The values for $K$ reported here were chosen
    so that the generated examples would fit a 
    predetermined
    $L_2$ cut-off.
    }
    \vspace{-10pt}
    \def\arraystretch{1.1}
    \resizebox{\columnwidth}{!}{
    \begin{tabular}{|c|c|c|c|}
        \hline
        \textbf{Dataset}  & 
        \makecell{\textbf{$L_2$} \\  \textbf{cut-off}} &
         \makecell{\textbf{Working confidence} \\  \textbf{values} $(K)$} 
        & \makecell{\textbf{Examples for} \\ \textbf{each $K$ $(n)$}} 
        \\\hline
        \MNIST   & 3.0 & $\{0, 5, 10, 15\}$ & 3000  \\\hline
        \FASHION & 1.0 & $\{0, 20, 40, 60\}$ & 3000  \\\hline
        \CIFAR   & 1.0 & $\{0, 20, 40, 60, 80, 100\}$ & 2000  \\\hline
    \end{tabular}
       }\label{table:para-adv}
    \vspace{-0.5em}
\end{table}

Appendix~\ref{app:training} provides additional details on training for each defense scheme.

\input{src/table_whitebox_adv_l2.tex}

\subsection{White-box Evaluation}
\label{eval-subsect:white-box-i}

We first evaluate the effectiveness of the defense mechanisms under white-box attacks.  We 
apply the \CW white-box attack with confidence 0 to generate targeted adversarial examples, 
and measure the $L_2$. distance of the generated adversarial examples.  
We consider both  single-model attack, where the adversarial example targets a single model, and
multi-8 attack, where the adversarial example aims at attacking 8 similarly trained model at the
same time. This can be considered as a form of ensemble white-box attack~\cite{liu_CLS_16}.

Tables~\ref{table:wbox_single} and \ref{table:wbox_multi8} present the average 
$L_2$ distances of the generated examples for those generated adversarial examples. 
\RS{1}-Adv results in models that are more difficult to attack, requires on average the highest 
perturbations (measured in $L_2$ distance) among all evaluated defenses.  
Comparing to other methods, adversarial examples generated by RS-1 and RS-1-Dropout have either higher or comparable amount of distortion.
These again suggest RS offers additional protection against 
adversarial examples.

\input{src/figure/noise.tex}

\subsection{Model Stability}
\label{eval-subsect:model-stability}

Given a benign image and its variants with added noise, a more robust model should intuitively be 
able to tolerate a higher level of noise without changing its prediction results. We refer to this property as model stability. 
Here we evaluate whether models from a defense mechanism can correct label instances that are perturbed.  
This serves several purposes.  
First, in~\cite{gilmer_fcc_2019}, it is suggested that vulnerability to adversarial examples and low performance on 
randomly corrupted images, such as images with additive Gaussian noise, are two manifestations of the same 
underlying phenomenon.  Hence it is suggested that adversary defenses should consider robustness under such 
perturbations, as robustness under such perturbations are also indications of resistance against adversary attacks. 
Second, evaluating stability is identified in~\cite{carlini_apbrtgmk_2019, athalye_cw_2018} as a way to check whether 
a defense relies on obfuscated gradients to achieve its defense.  For such a defense, random perturbation may discover 
adversarial examples when optimized search based on gradients fail.  Third, some defense mechanisms (such as Magnet) rely 
on detecting whether an instance belongs to the same distribution as the training set, and consider an instance to be an 
adversarial example if it does.  However, when an input instance goes through some transformation that has little impact 
on human visual detection (such as JPEG compression), it will be considered as an adversarial example by the defense.  
This will impact accuracy of deployed systems, as the encountered instances may not always follow the training distribution.

\subsubsection{Stability with Added Gaussian Noise}
We measure how many predictions would change if a certain amount of Gaussian noise is introduced
to a set of benign images.  
For a given dataset and a model, we use the first $1,000$ images from the test dataset. 
We first make a prediction on those 
selected images and store the results as \textit{reference predictions}. 
Then, for each selected image and chosen $L_2$ distance, 
we sample Gaussian noise, scale it to the desired $L_2$ value, and
add the noise to the image. Pixel values are clipped if necessary,
to make sure the new noisy variant is a valid image. 
We repeat this process 20 times (noise sampled independently per iteration). 

Fig.~\ref{fig:noise_stable} shows the effect of Gaussian noise on prediction stability 
for each training method (averaged over the 16 models trained in 
Sec.~\ref{eval-subsect:baseline}). 
Model stability inevitably drops for each scheme as the amount of Gaussian noise as measured by $L_2$ increases.  
However, different schemes behave differently when $L_2$ increases.  

For MNIST, most schemes have stability above 99\%, even when $L_2$ is as large as 5.
However, Magnet has 
stability approaching 0 when the $L_2$ distance is greater than 1, because majority 
of those instances are rejected by Magnet.

For Fashion-MNIST, we see more interesting differences among the schemes.  The two approaches that have highest stability are the two with adversarial training.  When $L_2=2.5$, RS-1-ADV has stability 87.4\%, and ADV has stability 86\%.  Other schemes have stability around 60\%; among them, RS-1 and RSD-1 have slightly higher stability than others. 

For CIFAR-10, we see that RS-1-ADV, RSD-1, and RS-1 have the highest stability as the amount of noise increases.  When $L_2=2.5$, they have stability 87.9\%, 81.7\%, 83\%, respectively.  The other schemes have stability 70\% or lower.  

Furthermore, on all datasets, RS-1-ADV, RSD-1, and RS-1 give consistent results.  
Recall that we trained 16 models for each scheme,  Fig.~\ref{fig:noise_stable} also plots the standard deviation of the stability result of the 16 models.  RS-1-ADV, RSD-1, and RS-1 have very low standard deviation, which in turn also suggest more consistent behavior when facing perturbed images.

\subsubsection{Stability with JPEG compression}
Given a set of benign images, we measure how many predictions would change if JPEG compression is applied to images. 
For a given test dataset and a model, we compare the prediction on 
the benign test dataset (\textit{reference predictions}) with the 
prediction on JPEG compressed test dataset with a fixed chosen JPEG 
compression quality (\textbf{JCQ}).
For the sake of time efficiency, for this particular set of experiments, we 
reduced the number of iterations used by RC to one-tenth of its original algorithm. 

Fig.~\ref{fig:jpeg_stable} shows the effect of JPEG compression on prediction stability 
for each training method (averaged over the 16 models trained). 
Model stability decreases for each scheme as the JCQ (ranges $10-100$) decreases. 

For \MNIST, most schemes achieve stability over 99, even if the JCQ is 10. Magnet is 
the outlier, which has a stability of around 50 when the JCQ is 70, and has a stability 
of less than 20 when the JCQ is less than or equal to 40, because of the high 
rejection rate of MagNet. 
We believe that both of these results are related to the fact that 
MNIST images have black backgrounds that span most of the image.   Noises introduced by JPEG compression result mostly in 
perturbations in the background that are ignored by most NN models. Since Magnet uses autoencoders to detect deviations from the input 
distributions, these noises trigger detection.  
Since Magnet aims at detecting perturbed images, this should not be 
considered as a weakness of Magnet.

For \FASHION, we see that RS-1-ADV, RSD-1, and RS-1 outperform other schemes on 
the stability as the JCQ decreases. When JCQ $=10$, they have stability 85.2\%, 
80.9\%, 84.4\%, respectively. The other schemes have stability 80\% or lower; The 
closest to the RS-class among other schemes is ADV.

For \CIFAR, we see that RS-1-ADV, RSD-1, and RS-1 have the highest stability as the JCQ decreases.  When JCQ $=10$, they have stability 60.9\%, 55.6\%, 55.4\%, respectively.  The other schemes have stability 50\% or lower; the highest among the other schemes is RC.

\input{src/figure/strategy_all}

\subsection{Evaluating Attack Strategies}
\label{eval-subsect:attk-strategy}

Here we empirically show that our proposed attack strategy, as presented in 
Sec.~\ref{sect:attacker-strategy}, can indeed generate adversarial examples that 
are more transferable. 
In attacks like the \CW attack, a higher 
confidence value will typically lead to more transferable examples, but the amount 
of perturbation would usually increase as well, sometimes making the example 
noticeably different under human perception.

Intuitively, a better attack strategy should give more transferable adversarial examples 
using less amount of distortion. Hence we use \textit{Distortion vs Transferability} to 
compare $3$ possible attack strategies. Similar to previous experiments, we 
measure the amount of distortion using $L_2$ distance. 
In Fig.~\ref{fig:attk_strategy} we present the effectiveness of each attack strategy, averaged across the 9 schemes. 

The first strategy we evaluated is a standard \CW attack which generates adversarial 
examples using only one surrogate model, dubbed \emph{`Single'}. Recall that for
each training/defense method, we have $16$ models that are surrogates of each other 
(Sec.~\ref{eval-subsect:baseline}). For each surrogate model, we randomly select 
half of the original dataset as the training dataset, since the adversary may not have 
full knowledge of the training dataset under the transfer attack setting. For the 
\emph{Single} 
strategy, we apply the \CW attack on $4$ of the models independently to generate a 
pool of adversarial examples. The transferability of those examples are then 
measured and averaged on the remaining $12$ target models.
Regardless of the training methods and defense 
mechanisms in place, adversarial examples generated using the \emph{Single} 
strategy often have limited transferability, especially when the allowed amount of 
distortion ($L_2$ distance) is small.

The second attack strategy that we evaluate is to 
generate adversarial examples using multiple surrogate models. 
For this, we 
use $8$ of the $16$ surrogate models for generating attack examples.
The \CW attack can be adapted to handle this case with a slightly different loss 
function. In our experiments, we use the sum of the loss functions of the $8$ surrogate 
models as the new loss function. 
We also use slightly lower confidence values than in 
Sec.~\ref{eval-subsect:adv-examples}
($\{0,10,20,30\}$ for \FASHION, 
$\{0,20,40,60\}$ \CIFAR).
The transferability of the generated adversarial examples are then measured and 
averaged on the 
remaining $8$ models as the target. We refer to this as \emph{`Multi 8'}. 
As shown in Fig.~\ref{fig:attk_strategy}, 
given the same limit on the amount of 
distortion ($L_2$ distance), a significantly higher percentage of 
examples generated using the \emph{Multi 8} strategy are transferable than those 
found using the \emph{Single} strategy.

Additionally, we evaluate a third attack strategy that is based on \emph{Multi 8}. As 
discussed in Sec.~\ref{sect:attacker-strategy}, given enough surrogate models, one 
can further use some of them for validating adversarial examples. 
For those adversarial examples generated by  the \emph{Multi 8} 
strategy, we keep them only if they can be transferred to at 
least $5$ of the $7$ validation models, hence we refer to this 
strategy \emph{Multi 8 \& Passing 5/7 Validation}.
The remaining model is used as the attack target, and we measure the 
transferability of examples that passed the \emph{5/7 Validation}.
For this attack strategy, the 
measurements shown in Fig.~\ref{fig:attk_strategy} is the average of $8$ rotations 
between target model and validation models. Comparing to \emph{Multi 8} and 
\emph{Single}, adversarial examples that passed the \emph{5/7 Validation} are 
significantly more likely to transfer to the target model, even when 
the amount of perturbation is small. 

This shows that simple strategies like \emph{Single} are indeed not realizing the full 
potential of a resourceful attacker, and
our proposed attack strategy of using multiple models for the generation and 
validation of adversarial examples is indeed superior. 
In the reset of this section, we will be using the most effective attack strategy 
of \emph{Multi 8 \& Passing 5/7 Validation}.

\subsection{Translucent-box Evaluation}
\label{eval-subsect:gray-box-attk}

Here we evaluate the effectiveness of different schemes based on the transferability 
of 
adversarial examples generated using the \emph{Multi 8 \& Passing 5/7 Validation} attack strategy. 

The results of our translucent-box evaluation are shown in 
Fig.~\ref{fig:heatmap-greyscale}.  
Adversarial examples are grouped into buckets based on their $L_2$ 
distance. For each bucket, we use grayscale to indicate the average validation passing rate for each scheme. 
Passing rate from 0\% to 100\% are mapped to pixel value from 0 to 255 in a linear scale.  
There are four rows, each correspond to adversarial examples with a certain $L_2$ 
range.  
Each column illustrates to what extent a target defense scheme resist adversarial examples generated from attacking different methods.

Examining the columns for Standard and Dropout, we can see that Standard and Dropout are in general most vulnerable.  
Distillation and RC are almost equally vulnerable.  Magnet can often resist adversarial examples generated by targeting other defenses, but are vulnerable to ones generated specifically targeting it.  

\input{src/cross/transfer_cross_all.tex}

Overall, across the three datasets, RS-1-Adv performs the best, and is significantly better than Dropout-Adv. 
This suggests that Random Spiking offers additional protection against adversarial examples.  
RS-1 and RS-1-Dropout also perform consistently well across the three datasets.
RC performs noticeably well on \MNIST and \FASHION, likely because the images 
were all in 8-bit grayscale, and its advantages diminish on \CIFAR which contains 
images of 24-bit color.

%% file: src/table_dataset_info.tex
\begin{table}
    \centering
    \caption{Overview of datasets}
    \vspace{-1em}
    \def\arraystretch{1.1}
    \setcellgapes{0pt}
    \makegapedcells
    \resizebox{\columnwidth}{!}{
    \begin{tabular}{|c|c|c|c|c|}
        \hline
        \textbf{Dataset}  & 
        \makecell{\textbf{Image size}} &
         \makecell{\textbf{Training} \\  \textbf{Instances}} 
        & \makecell{\textbf{Test} \\  \textbf{Instances}} 
        & \makecell{\textbf{Color} \\ \textbf{space}}
        \\\hline
        \MNIST   & $28\times28$ & 60,000 & 10,000 & 8-bits Gray-scale \\\hline
        \FASHION & $28\times28$ & 60,000 & 10,000 & 8-bits Gray-scale \\\hline
        \CIFAR   & $32\times32$ & 50,000 & 10,000 & 24-bits True-Color\\\hline
    \end{tabular}
       }\label{table:dataset-info}
    \vspace{-1em}
\end{table}

%% file: src/test-error-table-full-data.tex
\begin{table}
    \caption{Test errors (mean$\pm$std).
    }
    \vspace*{-10pt}
    \centering
    \def\arraystretch{1.1}
    \setlength{\tabcolsep}{4pt}    
    \label{table:model-acc-full}
    \resizebox{\columnwidth}{!}{
    \begin{tabular}{l|l|c|c|c}
        \Xhline{1pt}
        \multicolumn{2}{l|}{ }                             &  \MNIST & \FASHION & \CIFAR 
        \\\hline\hline
        Standard                       & Single pred. & $0.77\pm0.05\%$ & 
        $4.94\pm0.19\%$ & $4.38\pm0.21\%$ \\\hline
        
        \multirow{1}{*}{Dropout}
                                        & MC Avg.            & $0.67\pm0.07\%$ & $4.75\pm0.09\%$ &
                                        $4.46\pm0.25\%$ \\\hline

        \multirow{1}{*}{Distillation}
                                        & MC Avg.           & $0.78\pm0.05\%$ & $4.81\pm0.18\%$ &
                                        $4.33\pm0.27\%$ \\\hline

        \multirow{1}{*}{\RS{1}}
                                        & MC Avg.            & $0.88\pm0.09\%$ & $5.34\pm0.10\%$ &
                                        $5.59\pm0.22\%$ \\\hline
        
        \multirow{1}{*}{\RS{1}-Dropout}
                                        & MC Avg.            & $0.71\pm0.07\%$ & $5.32\pm0.17\%$ &
                                        $5.81\pm0.27\%$ \\\hline
 
        \multirow{1}{*}{\RS{1}-Adv}
                                        & MC Avg.            & $0.98\pm0.11\%$ & $5.49\pm0.16\%$ &
                                        $6.20\pm0.40\%$ \\\hline

        \multirow{2}{*}{Magnet}        & \textit{Det. Thrs.}          & $0.001$        & 
        $0.004$        
        & $0.004$        \\ \cdashline{2-5}[1.5pt/2pt]
                                        & MC Avg.            & $0.87\pm0.06\%$ & $5.36\pm0.17\%$ &
                                        $5.52\pm0.24\%$ \\\hline

        \multirow{1}{*}{Dropout-Adv}
                                        & MC Avg.            & $0.69\pm0.07\%$ & $4.76\pm0.11\%$ &
                                        $4.71\pm0.19\%$\\\hline

        \multirow{2}{*}{RC}            & \textit{$L_2$ noise}       & $0.4$           & 
        $0.02$          & $0.02$          \\ \cdashline{2-5}[1.5pt/2pt]
                                       & Voting            & $0.77\pm0.11\%$ & $5.39\pm0.23\%$ & 
                                       $5.72\pm0.46\%$ \\
        \Xcline{1-5}{1pt}
    \end{tabular}
    }
    \vspace*{-0.75em}
\end{table}

%% file: src/table_whitebox_adv_l2.tex
\begin{table}
    \caption{\CW targeted Adv Examples $L2$ (mean$\pm$std) when attacking a single model. \label{table:wbox_single}}
    \vspace*{-10pt}
    \centering
    \def\arraystretch{1.1}
    \setlength{\tabcolsep}{8pt}    
    \label{table:adv-l2-full-single}
    \resizebox{\columnwidth}{!}{
    \begin{tabular}{lccc}
        \Xhline{1pt}
        &  \MNIST & \FASHION & \CIFAR \Tstrut\Bstrut\\\hline
        Standard       & $2.12 \pm 0.69$ & $0.12 \pm 0.08$ & $0.17 \pm 0.08$ \\
        Dropout        & $1.80 \pm 0.52$ & $0.14 \pm 0.07$ & $0.17 \pm 0.08$ \\
        Distillation   & $2.02 \pm 0.63$ & $0.13 \pm 0.07$ & $0.17 \pm 0.07$ \\
        \RS{1}         & $2.06 \pm 0.76$ & $0.31 \pm 0.16$ & $0.32 \pm 0.14$ \\
        \RS{1}-Dropout & $1.79 \pm 0.86$ & $0.36 \pm 0.21$ & $0.32 \pm 0.15$ \\
        \RS{1}-Adv     & $2.36 \pm 0.80$ & $0.56 \pm 0.30$ & $0.39 \pm 0.18$ \\
        Magnet         & $2.22 \pm 0.65$ & $0.28 \pm 0.15$ & $0.29 \pm 0.21$ \\
        Dropout-Adv    & $2.44 \pm 0.66$ & $0.33 \pm 0.15$ & $0.18 \pm 0.07$ \Bstrut\\
        \Xhline{1pt}
    \end{tabular}
    }
    \vspace{-0.75em}
\end{table}

\begin{table}
    \caption{
    \CW Adv Examples $L2$ (mean$\pm$std) with \emph{Multi 8} attack 
    strategy. \label{table:wbox_multi8}}
    \vspace*{-1em}
    \centering
    \def\arraystretch{1.1}
    \setlength{\tabcolsep}{8pt}    
    \resizebox{\columnwidth}{!}{
    \begin{tabular}{lccc}
        \Xhline{1pt}
        &  \MNIST & \FASHION & \CIFAR \Tstrut\Bstrut\\\hline
        Standard       & $2.50 \pm 0.77$ & $0.22 \pm 0.15$ & $0.25 \pm 0.10$ \\
        Dropout        & $2.29 \pm 0.65$ & $0.25 \pm 0.13$ & $0.26 \pm 0.10$ \\
        Distillation   & $2.37 \pm 0.71$ & $0.24 \pm 0.14$ & $0.33 \pm 0.13$ \\
        \RS{1}         & $2.77 \pm 0.82$ & $0.54 \pm 0.25$ & $0.49 \pm 0.18$ \\
        \RS{1}-Dropout & $2.77 \pm 0.93$ & $0.61 \pm 0.30$ & $0.51 \pm 0.18$ \\
        \RS{1}-Adv     & $3.18 \pm 0.88$ & $1.04 \pm 0.44$ & $0.64 \pm 0.23$ \\
        Magnet         & $2.68 \pm 0.75$ & $0.54 \pm 0.25$ & $0.47 \pm 0.24$ \\
        Dropout-Adv    & $2.93 \pm 0.70$ & $0.57 \pm 0.23$ & $0.29 \pm 0.10$ \Bstrut\\
        \Xhline{1pt}
    \end{tabular}
    }
    \vspace{-0.75em}
\end{table}

%% file: src/figure/noise.tex
\begingroup
\begin{figure*}
    \begin{subfigure}{\textwidth}
        \centering
        \includegraphics[width=0.90\textwidth]{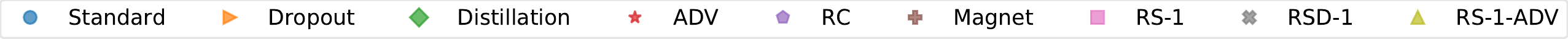}
    \end{subfigure}
    \vspace{-0.5em}\\
    \foreach \setname/\setsymbl in {mnist/\MNIST, fashion/\FASHION, cifar/\CIFAR} {
        \begin{subfigure}{0.30\textwidth}
            \caption{Prediction Stability (\setsymbl)}
            \vspace{0pt}
            \includegraphics[width=0.95\textwidth]{./src/figure/noise_full_l2/\setname_plot_unchg.pdf}
            \label{fig:noise_unchg_guassian:\setname}
        \end{subfigure}
    }    \vspace{-12pt}
    \caption{Evaluating model stability with Gaussian Noise}
    \label{fig:noise_stable}
    \foreach \setname/\setsymbl in {mnist/\MNIST, fashion/\FASHION, cifar/\CIFAR} {
        \begin{subfigure}{0.30\textwidth}
            \caption{Prediction Stability (\setsymbl)}
            \vspace{0pt}
            \includegraphics[width=0.95\textwidth]{./src/figure/jpeg_full/\setname_plot_unchg.pdf}
            \label{fig:noise_unchg_jpeg:\setname}
        \end{subfigure}
    }    \vspace{-1em}
    \caption{Evaluating model stability with JPEG compression}
    \label{fig:jpeg_stable}
    \vspace{-1em}
\end{figure*}
\endgroup

%% file: src/figure/strategy_all.tex
\begingroup
\begin{figure*}
    \centering
    \begin{subfigure}{0.93\textwidth}
        \centering
        \includegraphics[width=0.5\textwidth]{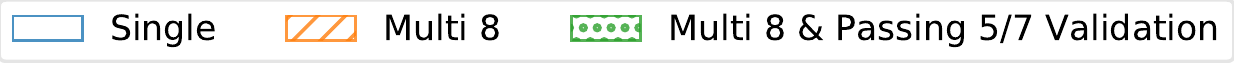}
    \end{subfigure}
    \vspace{0.0pt}\\
    \foreach \dname in {\MNIST, \FASHION, \CIFAR}{
        \begin{subfigure}{0.32\textwidth}
            \centering
            \small\qquad\dname
        \end{subfigure}
    }\vspace{0.5pt}\\
    \foreach \evalname/\evelstname/\setimgs in {
        all/Standard/{mnist, fashion, cifar}} {
        \foreach \curimg in \setimgs {
            \begin{subfigure}{0.32\textwidth}
                \ifx\curimg\emptytext
                    \begin{tikzpicture}
                        \node[inner sep=15pt,outer sep=0pt] (A) at (0,0) { };
                        \node[inner sep=0pt,outer sep=0pt] (B) at (5,0.8) { };
                        \draw (A) edge[-] (B);
                    \end{tikzpicture}
                \else
                    \includegraphics[width=0.95\textwidth]{./src/figure/strategy/\curimg/\evalname-l2_mn.pdf}
                \fi
            \end{subfigure}
        }\\
    }
    \vspace{-1.25em}
    \caption{Transferability of adversarial examples found by the 3 attack strategies}
    \label{fig:attk_strategy}
    \vspace{-1em}
\end{figure*}
\endgroup

%% file: src/cross/transfer_cross_all.tex
\begingroup
\begin{figure*}
    \foreach \dname/\dddname in {\MNIST/mnist, \FASHION/fashion, \CIFAR/cifar}{
        \begin{subfigure}{0.30\textwidth}
            \caption{\dname}
            \includegraphics[width=0.95\textwidth]{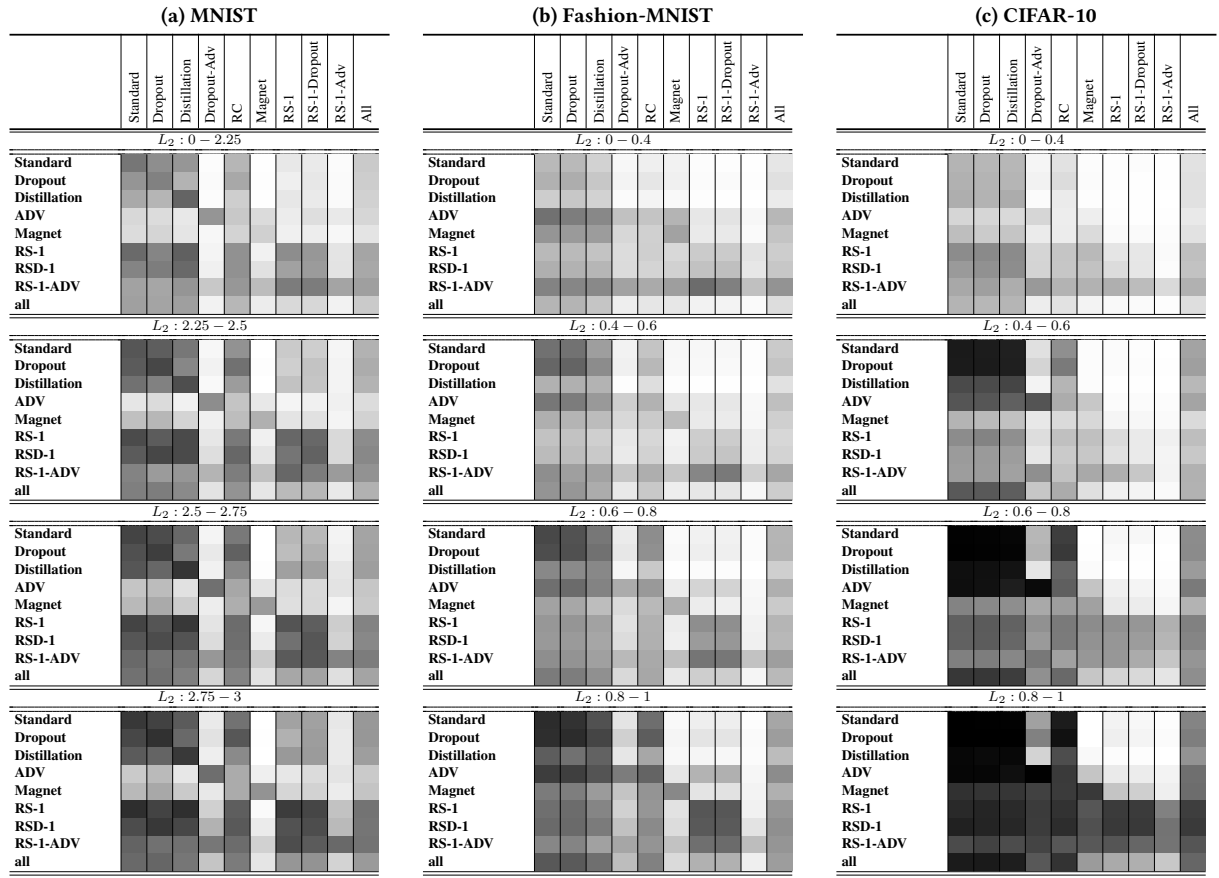}
            
            \label{fig:5-7-valid:\dddname}
        \end{subfigure}
            }\vspace{-0.5em}
    \caption{Average passing rate of 5/7 validation }
                    This heat map shows the resilience of each scheme against the adversarial examples generated from all schemes, under a fixed allowance of $L_2$. 
                    Each column illustrates to what extent a target defense scheme resists adversarial examples generated from attacking different methods. 
    \label{fig:heatmap-greyscale}
    \vspace{-0.25em}
\end{figure*}
\endgroup

%% file: src/6_related.tex
\label{sec:related}

\mypara{Other attack algorithms.} 
There are other attack algorithms such as JSMA \cite{papernot_MJFCS_15}, FGS
\cite{goodfellow_fgd_15}, PGD\cite{madry2017towards}, and DeepFool \cite{moosavi_Dezfooli_15}.  The general consensus seems 
to be that the \CW attack is the current state-of-the-art \cite{carlini_oakland_17,carliniw_advdector_17,chen_SZYH_2017}. 
Though our evaluation results are based on the \CW attack,
our evaluation framework
is not tied to a particular attack and can use other algorithms.  

\mypara{Other defense mechanisms.} Some other defense
mechanisms have been proposed
\cite{papernot_MWJS_15, mengccs17, cao_G_2017, xu_EQ_2018, xie_wzry_2018, xie_wvyh_2019_CVPR} 
in the literature. For example,  Xu
\etal.~proposed to use feature squeezing techniques
such as color bit depths reduction and pixel spatial smoothing
to detect adversarial examples~\cite{xu_EQ_2018}.
Xie \etal.~proposed to use Feature Denoising \cite{xie_wvyh_2019_CVPR}
to improve the robustness of the Neural Network Model.
Due to limit in time and space, we selected representative methods from each broad class (\eg, MagNet for the detection approach). 
We leave comparison with other mechanisms as future work.   

\mypara{Beyond images.} Other research efforts have explored
possible attacks against neural network models specialized for other
purposes, for example, speech to text \cite{carlini_W_2018}.  
We focus on images, although the evaluation methodology and the idea of random spiking should be applicable to
these other domains. 

\mypara{Alternative similarity metrics.} Some researchers have argued that $L_p$
norms insufficiently capture human perception, and have proposed alternative
similarity metrics like SSIM \cite{sharif_br_2018}. It is however, not immediately clear how to adapt
such metrics in the \CW attack. We leave further investigations on the impacts of
alternative similarity metrics on adversarial examples for future work.
\vspace{-0.5em}

%% file: src/7_conclusion.tex
\label{sec:conclusion}

In this paper, we present a careful analysis of possible adversarial models 
for studying the phenomenon of adversarial examples.  We propose an evaluation methodology 
that can better illustrate the strengths and limitations of different mechanisms.   
As part of the method, we introduce a more powerful and meaningful adversary strategy. 
We also introduce Random Spiking, a randomized technique that generalizes dropout.  
We have conducted extensive evaluation of Random Spiking and several other defense mechanisms, 
and demonstrate that Random Spiking, especially when combined with adversarial training, offers 
better protection against adversarial examples when compared with other existing defenses.

\section*{Acknowledgements}  This work is supported by the Northrop Grumman 
Cybersecurity Research Consortium under a Grant titled ``Defenses Against Adversarial Examples'' 
and by the United States National Science Foundation under Grant No. 1640374.

%% file: src/appendices.tex
\section{Appendix}
\subsection{More Information on Training} \label{app:training}

Model architecture and training parameters are in Tables~\ref{table:model-arch} and 
~\ref{table:model-para}.  For \MNIST, we use the same as in MagNet~\cite{mengccs17, carlini_oakland_17}). 
For \FASHION and \CIFAR, they are  identical to the WRN~\cite{zagoruyko_K_16}. 
Test errors of the nine schemes are in Table~\ref{table:model-acc-full}. For each scheme and dataset, we train 16 models and report the mean and standard deviation. Additional information is provided below.

\input{src/architecture-and-params}

\mypara{Magnet.}
We use the trained \emph{Dropout} model as the prediction model, and train the Magnet defensive models 
(reformers and detectors)~\cite{mengccs17} based on the publicly released 
Magnet implementation\footnote{\url{https://github.com/Trevillie/MagNet}}.  
Identical to the settings\footnote{Regarding the Detector settings, a small discrepancy exists between the paper and the released source code. After confirming with the authors, we follow what is given by the source code.}
presented in the original Magnet paper~\cite{mengccs17},
for \MNIST, we use Reformer I, Detector I/$L_2$ and Detector II/$L_1$, with detection threshold set to $0.001$.
Since \FASHION was not studied in~\cite{mengccs17}, we use the same model 
architecture as \CIFAR presented in the original Magnet paper~\cite{mengccs17}.
For \FASHION and \CIFAR, we use Reformer II, Detector II/$L_1$, Detector II/$T10$ and Detector II/$T40$,
and with a detection threshold (rate of false positive) of $0.004$, which results in test error rates comparable to those of the other schemes.

\mypara{\randomspiking with standard model (RS-1).}
A \randomspiking (RS) layer is added after the first convolution layer in the \emph{standard} architecture. 
We choose $p=0.8$, so that $20\%$ of all neuron outputs are randomly spiked.  

\mypara{\randomspiking with \dropout (RS-1-Dropout).}
We add the RS layer to the \emph{Dropout} scheme. All other parameters are identical to what we 
used for \RS{1}. We also use \textbf{\emph{RSD-1}} as a shorthand to refer to this scheme.

\mypara{Distillation.}
We use the same network architecture and parameters as we did for the training of \emph{Dropout} models. 
Identical to the configuration used in \cite{carlini_oakland_17},
we train with temperature $T=100$ and test with $T=1$ for all three datasets.

\mypara{Region-based Classification (RC).}
We use the \emph{Dropout} models for \emph{RC}. 
For each test example, we generate $t$ additional examples, where for each pixel, a noise was randomly chosen from $(-r, r)$ and added to it.
Prediction is then made with majority voting on the $t$ input examples.
Identical to the original RC paper~\cite{cao_G_2017}, we use $t=10,000$ for \MNIST and $t=1,000$ for 
\CIFAR.  We also use $t=1,000$ for \FASHION. 
We choose values for $r$ ($r=0.4$ for \MNIST, and $r=0.02$ for \FASHION and \CIFAR) so that the test errors would be comparable to the other mechanisms.

\mypara{Adversarial Dropout (Dropout-Adv).}
To use adversarial training with \emph{\dropout}, we leverage the trained \dropout 
model from before as the target model for generating adversarial examples. 
We generated $12,000$ adversarial examples for each \emph{Dropout} model by perturbing training instances.  
To ensure that the adversarial examples indeed should be classified under the original label, 
we sort the adversarial examples according to their $L_2$ distances 
in ascending order, and add only the first $10,000$ examples into the training dataset. 
These examples have $L_2$ distances lower than the cutoff mentioned earlier. 
We then apply the \dropout training procedure as described before on the new training dataset.

\mypara{Adversarial \randomspiking (RS-1-Adv).}
For this adversarial training method, we use \emph{\RS{1}} as the target model. 
 The training parameters and procedure are largely identical to what were described 
 for Dropout-Adv above.

\subsection{Proof of Proposition~\ref{def:RSprop}}\label{app:proof}

\begin{proof}[Proof sketch]
The Monte Carlo sampling used in the RS neural network optimization
gives an unbiased estimate of the gradient
\vspace{-0.1em} 
\begin{align*}
&\sum_{\forall\boldsymbol{b}}\int_{\boldsymbol{\epsilon}}\frac{\partial}{\partial\boldsymbol{W}}L\left(y,\hat{y}(x,\boldsymbol{b},\boldsymbol{\epsilon},\boldsymbol{W})\right)p(\boldsymbol{b})f(\boldsymbol{\epsilon})d\boldsymbol{\epsilon}  \\
&=\frac{\partial}{\partial\boldsymbol{W}}\sum_{\forall\boldsymbol{b}}\int_{\boldsymbol{\epsilon}}L\left(y,\hat{y}(x,\boldsymbol{b},\boldsymbol{\epsilon},\boldsymbol{W})\right)p(\boldsymbol{b})f(\boldsymbol{\epsilon})d\boldsymbol{\epsilon},
\end{align*}
with the above equality given by the linearity of the expectation
and integral operators. That is, the RS neural network optimization
is a Robbins-Monro stochastic optimization~\cite{bottou2010large} that minimizes
\vspace{-0.1em} 
\[
\boldsymbol{W}^{\prime}=\arg\!\min_{\boldsymbol{W}}\sum_{\forall\boldsymbol{b}}\int_{\boldsymbol{\epsilon}}L(y,\overline{\hat{y}}(x,\boldsymbol{W}))p(\boldsymbol{b})f(\boldsymbol{\epsilon})d\boldsymbol{\epsilon}.
\]
$L$ is convex on $\hat{y}$, then by Jensen's inequality 
\vspace{-0.1em} 
\begin{align*}
&\sum_{\forall\boldsymbol{b}}\int_{\boldsymbol{\epsilon}}L\left(y,\hat{y}(x,\boldsymbol{b},\boldsymbol{\epsilon},\boldsymbol{W})\right)p(\boldsymbol{b})f(\boldsymbol{\epsilon})d\boldsymbol{\epsilon}\geq\\ 
&\quad L\left(y,\sum_{\forall\boldsymbol{b}}\int_{\boldsymbol{\epsilon}}\hat{y}(x,\boldsymbol{b},\boldsymbol{\epsilon},\boldsymbol{W})p(\boldsymbol{b})f(\boldsymbol{\epsilon})d\boldsymbol{\epsilon}\right)\equiv L\left(y,\overline{\hat{y}}(x,\boldsymbol{W})\right).
\end{align*}
Thus, the RS neural network minimizes an upper bound of the loss of
the ensemble RS model $\overline{\hat{y}}(x,\boldsymbol{W})$, yielding a proper variational inference procedure~\cite{blei2017variational}.
\end{proof}

%% file: src/architecture-and-params.tex
\begin{table}[bth]
    \caption{Mode Architectures. We use WRN-28-10 for \FASHION and \CIFAR ($k=10, N=4$). }
    \vspace*{-10pt}
    \centering
    \def\arraystretch{1.1}
    \setlength{\tabcolsep}{3pt}
    \label{table:model-arch}
    \resizebox{\columnwidth}{!}{    
    \begin{tabular}{ll||lll|ll}
        \Xhline{1pt}
        \MNIST                   &                     & &\multicolumn{2}{l|}{\FASHION} & 
        \multicolumn{2}{l}{\CIFAR}
        \\ \cline{1-7}
        & & 
        Group & 
        \makecell{Output\\ Size} 
        & Kernel, Feature & \makecell{Output\\ 
        Size} & Kernel, Feature     \\\cline{3-7}
        Conv.ReLU               & $3\times3\times32$   &  Conv1  & $28\times28$ & 
        $[3\times3, 16]$ & $32\times32$ & $[3\times3, 16]$    \Tstrut\\
        Conv.ReLU      & $3\times3\times32$ & \multirow{3}{*}{Conv2} & 
        \multirow{3}{*}{$28\times28$}
        & \multirow{3}{*}{$
            \begin{bmatrix}
            3\times3, 16\times k \\
            3\times3, 16\times k
            \end{bmatrix}\times N$
        }
        & \multirow{3}{*}{$32\times32$}
        & \multirow{3}{*}{$
            \begin{bmatrix}
            3\times3, 16\times k \\
            3\times3, 16\times k
            \end{bmatrix}\times N$
        }        \\
        Max Pooling & $2\times2$ & & & & &        \\
        Conv.ReLU & $3\times3\times64$ & & & & &        \\
        Conv.ReLU & $3\times3\times64$ &
        \multirow{3}{*}{Conv3} & 
        \multirow{3}{*}{$14\times14$}
        & \multirow{3}{*}{$
            \begin{bmatrix}
            3\times3, 32\times k \\
            3\times3, 32\times k
            \end{bmatrix}\times N$
        }
        & \multirow{3}{*}{$16\times16$}
        & \multirow{3}{*}{$
            \begin{bmatrix}
            3\times3, 32\times k \\
            3\times3, 32\times k
            \end{bmatrix}\times N$
        }        \\
        Max Pooling             & $2\times2$         & & & & &         \\
        Dense.ReLU              & $200$             &  & & & &         \\
        \multirow{2}{*}{Dense.ReLU}              & \multirow{2}{*}{$200$} & \multirow{2}{*}{Conv4} & 
        \multirow{2}{*}{$7\times7$}
        & \multirow{2}{*}{$
            \begin{bmatrix}
            3\times3, 64\times k \\
             3\times3, 64\times k
            \end{bmatrix}\times N$
        }
        & \multirow{2}{*}{$8\times8$}
        & \multirow{2}{*}{$
            \begin{bmatrix}
            3\times3, 64\times k \\
            3\times3, 64\times k
            \end{bmatrix}\times N$
        }        \\
                      &              & & & & &        \\
        Softmax                 & $10$              &Softmax  & $10$ & & $10$ &        \\
        \Xcline{1-7}{1pt}
    \end{tabular}
    }
    \vspace{-1em}
\end{table}

\begin{table}[tbh]
    \caption{Training Parameters.}
    \vspace*{-10pt}
    \centering
    \label{table:model-para}
    \resizebox{\columnwidth}{!}{
        \begin{tabular}{lll}
            \Xhline{1pt}
            Parameters               & \MNIST & \FASHION \& \CIFAR \Tstrut\Bstrut       \\ \hline
            Optimization Method      & SGD   & SGD                        \Tstrut \\
            Learning Rate            & 0.01  & \multirow{2}{*}{\shortstack[l]{0.1 initial, 
            multiply by 0.2
                    \\ at 60, 120 and 160 epochs}} \\
            &       &                            \\
            Momentum                 & 0.9   & 0.9                        \\
            Batch Size               & 128   & 128                        \\
            Epochs                   & 50    & 200                        \\
            Dropout (Optional)       & 0.5   & 0.1                        \\
            Data Augmentation        & -     & \multirow{3}{*}{\shortstack[l]{
                    \FASHION: Shifting + Horizontal Flip\\
                    \CIFAR: Shifting + Rotation + \\
                    Horizontal Flip + Zooming + Shear
                }} \Bstrut\\
                &       &\\
                &       &\\
                \Xhline{1pt}             &
        \end{tabular}
    }
    \vspace{-1em}
\end{table}